# An Evolution of CNN Object Classifiers on Low-Resolution Images

Md. Mohsin Kabir
*Department of Computer Science and Engineering*
*Bangladesh University of Business & Technology*
Dhaka, Bangladesh
m97kabir@gmail.com

Abu Quwsar Ohi
*Department of Computer Science and Engineering*
*Bangladesh University of Business & Technology*
Dhaka, Bangladesh
quwsarohi@bubt.edu.bd

Md. Saifur Rahman
*Department of Computer Science and Engineering*
*Bangladesh University of Business & Technology*
Dhaka, Bangladesh
saifurs@gmail.com

M. F. Mridha
*Department of Computer Science and Engineering*
*Bangladesh University of Business & Technology*
Dhaka, Bangladesh
firoz@bubt.edu.bd

*Abstract*—Object classification is a significant task in computer vision. It has become an effective research area as an important aspect of image processing and the building block of image localization, detection, and scene parsing. Object classification from low-quality images is difficult for the variance of object colors, aspect ratios, and cluttered backgrounds. The field of object classification has seen remarkable advancements, with the development of deep convolutional neural networks (DCNNs). Deep neural networks have been demonstrated as very powerful systems for facing the challenge of object classification from high-resolution images, but deploying such object classification networks on the embedded device remains challenging due to the high computational and memory requirements. Using high-quality images often causes high computational and memory complexity, whereas low-quality images can solve this issue. Hence, in this paper, we investigate an optimal architecture that accurately classifies low-quality images using DCNNs architectures. To validate different baselines on low-quality images, we perform experiments using webcam captured image datasets of 10 different objects. In this research work, we evaluate the proposed architecture by implementing popular CNN architectures. The experimental results validate that the MobileNet architecture delivers better than most of the available CNN architectures for low-resolution webcam image datasets.

*Keywords— Deep learning, Computer vision, Convolutional neural networks, Image processing, Object classification.*

## I. INTRODUCTION

The amount of image data on the web has increased rapidly, with the internet and multimedia terminals' advancements. To solve large-scale visual problems using this huge amount of data, automatic data processing is anticipated with the help of computers. Image processing technology becomes more significant for accurate detection and classification of the object [1]. Nowadays, in image classification systems, people expect specific semantic categories of the object and the location in the image [2, 3]. Using the theories and algorithms of image processing and pattern recognition, the object detection approach detects the target object and determines the semantic categories, also marks the specific position of the intended object in the image [4, 5].

Deep learning approaches have been introduced in the object classification domain and have achieved vast success due to robustness. DL researchers have introduced convolutional neural networks (CNNs) that solve the pattern recognition problem linked with images. The transfer learning strategy has further improved the development of deep CNN-based architectures. The CNN model is initially trained on a comparatively large dataset in a transfer learning approach. The trained model is referred to as a pre-trained model, and the pre-trained model further identifies similar image patterns from the same or different domain of datasets. A transfer learning approach often helps to avoid the overfitting of deep learning methods on small datasets.

Several deep learning techniques are introduced for image classification using high-resolution images. But high-resolution images create more memory and computational complexity that is not suitable for an embedded device. Besides, low-quality webcam images contain skewness, and introduced models struggle to classify objects accurately. Thus, in this research work, we exploit the effectiveness of the popular DCNN architectures using low-quality webcam images. The overall contribution of the research can be summarized as,

• We investigate the existing CNN architectures that achieve better accuracy on small low-quality datasets. This is the first research endeavor that performs benchmarks on low-quality image classification.

• We experimented with six popular image recognition baseline strategies that include Densenet, Inception, Mobilenet, ResNet, VGG, and Xception and found that MobileNet gives better accuracy for low-quality webcam images.

The rest of this paper is constructed as follows: The related work is discussed in Section 2. The overall architecture of deep CNN is described in section 3. Section 4 describes the model's evaluation and compares the results of the architectures. At last, Section 5 concludes the paper.

## II. RELATED WORK

In computer vision, object classification in images and videos is one of the most primary issues. Deep learning architectures have been extensively conducted for numerous computer vision tasks, and recognition and detection of an object are recognized due to the power of learning peculiarity assertion. [25,26].

Industrial applications [10, 11], assisted driving [6, 8, 9] and video surveillance [7, 12, 13] are the basic area of applications of the object classification and detection system. Recently, for the embedded device, inquiring shallow deep neural network structures for object classification tasks is extensively increased.

Classical object classifiers use a sliding window algorithm to localize objects at diverse scales and aspect ratios. Real-time object classifiers are obtainable for particular categories, such as faces or movements [23, 24]. Many renowned object classification systems used cascade classifiers to classify an object, where instead of using a single classifier, cascades of more complex classifiers are used to the image [35, 36]. All cascade models use naive classifiers to reject big portions of the image. Elad et al. [38] conducts a features-based method on pixel values and extracts classifiers that increase the rejection rate. Romdhani et al. [39] construct a support vector machine method and then investigate the SVM with an order of support vector classifiers that usage non-linear optimization techniques. Keren et al. [37] proposed an anti-face classifier that simulates a normal distribution in the background of the input object.

Convolutional neural networks help to extract and learn features directly from a training dataset [18, 19]. To explore the basic characteristics of these features that are robust to different poses, rotations, scales, illumination, and camera qualities is a great challenge for any object classification system. Several approaches have been introduced to solve this difficulty [14, 15]. In convolutional neural networks, these object classifiers can be trained and learned simultaneously with the respective features [20, 21] or using excessive machine learning algorithms [16, 17, 22].

These proposed architectures do not fit on the embedded device due to the vast number of trainable parameters come from high-quality images. Besides, the skewness of low-quality webcam images struggles with the architecture to classify objects accurately. Thus, in this work, we evaluate famous CNN baseline architectures that used low-resolution webcam images and find out the best-fitted architecture. We evaluate six CNN baseline architecture and measure the accuracy using low-resolution webcam images and compare them with the result come from the high-quality image.

### III. METHODOLOGY

Different CNN architectures are implemented and benchmarked to perform object classification. In the following sub-sections, the general process and layers of various CNN architectures are briefed.

#### A. Data Pre-processing

Data normalization is a significant technique that assures that each input parameter has an identical data distribution. So each input image is reshaped into 64 by 64 pixels. While training the convolutional network, data normalization makes convergence faster. Therefore, each channel of the reshaped input images is normalized as,

$$Normalize(D) = \frac{\begin{bmatrix} d_{11} & \cdots & d_{1m} \\ \vdots & \ddots & \vdots \\ d_{n1} & \cdots & d_{nm} \end{bmatrix}}{127.5} \quad (1)$$

Where D is the single-channel object image matrix, n is the number of rows, and m is the number of columns of the object image matrix.

#### B. Baseline Architecture

A convolutional neural network is a deep learning algorithm that shares weight, add bias, downsample the input, and have local connection approaches that significantly shorten the constituent parameters and the complexity of the neural network architecture. CNN's can extract more extensive and abstract features than other approaches, such as Scale-Invariant Feature Transform, and Histogram of Oriented Gradient methods. By increasing the notion of the receptive field and sharing the weights and biases CNN reduces both the parameters of training and the complexity of the architecture. The convolutional neural networks are mainly developed by a series of convolutional layers, pooling layers, and fully connected or dense layers.

In this paper, we focus on evaluating benchmarked CNN architecture for dynamic object classification. Fig. 1 shows the CNN architecture, with the input layer (images of the object), convolutional layers, dense layer, and an output layer.

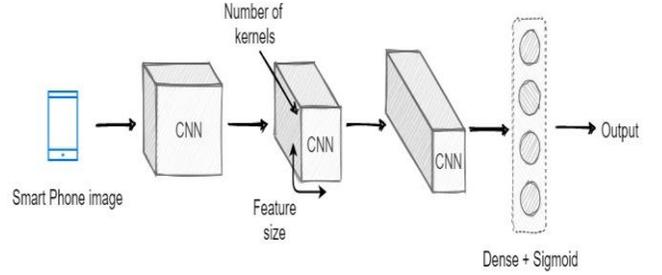

Fig. 1. The neural network architecture of object classification from low-quality image datasets. Each of the cubes represents an output of the convolution. The height and width are the gained information, and each cube's depth is equal to the number of kernels. Each convolution is followed by batch normalization and an activation layer. After the final convolution, it is converted into a linear set of nodes. Each node flows values through a sigmoid activation function.

*Convolutional Layers:* The convolutional layer is a fundamental part of convolutional neural network architecture. Using input images or feature maps the function of the convolutional layer extracts the features. Every layer consists of multiple convolution kernels, that helps to get multiple feature maps. The convolution procedure can be defined mathematically as follows:

$$x_j^l = f(\sum_{i \in M_j} x_i^{l-1} * k_{ij}^l + b_j^l) \quad (2)$$

Where $x_i^{l-1}$ is the characteristic map of the output of the previous layer, $x_j^l$ is the output of the ith channel of the jth convolution layer, M_j is a subset of the input feature maps used to calculate $u_j^l$, f(·) is called the activation function. Here, $k_{ij}^l$ is a convolution kernel, and $b_j^l$ is the corresponding offset.

*Batch Normalization:* Batch normalization is a technique that allows each layer of the model to train more independently.

To normalize the output of the previous layers, batch normalization is used. Having a slight regularization effect, it decreases the overfitting. It enhances the stability of deep neural network architectures and leads to quick convergence. The equation of normalization can be expressed as,

$$x'^{(k)}_i = \frac{x^{(k)}_i - \mu^{(k)}_B}{\sqrt{\sigma^{(k)2}_B}} \qquad (3)$$

Where $x'^{(k)}_i$ is the normalized value of the kth hidden unit. $\mu^{(k)}_B$ is the mean value, and $\sigma^{(k)2}_B$ is the variance of the kth hidden unit. B defines the data of a particular batch.

ReLu: ReLU function is used in every convolution layer for an easy calculation. If the Relu activation function obtains any non-positive value, it returns zero but for any positive value of x, it returns that input value. So the function can be expressed as,

$$ReLU = \max(0, x) \qquad (4)$$

Dense Layer: A fully Connected Layer or dense layer is a simple feed-forward neural network that connects every neuron in one layer to every neuron in the following layer. The output obtained from the final pooling layer or convolution layer is used as an input of a dense layer, which is first flattened and then fed into the layer. Generally, a fully connected layer can be represented as,

$$d(x) = Activation(w^T x + b) \qquad (5)$$

Here, w = [$w_1, w_2, …, w_n$]$^T$ represents the weight vector of the dense layer, and b represents the bias value of the dense layer.

Sigmoid: Sigmoid activation is used for the dynamic object classification task. If it gets a real value as input, it outputs another value [0,1]. In our model, it classifies the object classes. It can be defined as follows,

$$\sigma(x) = \frac{1}{1+e^{-x}} \qquad (6)$$

Here, x represents the input of the sigmoid function. The sigmoid function is used as the final activation function described in (6) Every node of the fully connected layer has a sigmoid function for classifying the object shape.

*C. Loss Function*

To calculate the loss, we have used categorical cross-entropy in the architectures. The categorical cross-entropy loss function calculates the loss of an input that is stated below,

$$L_i = -\sum_j (t_{i,j} \log(p_{i,j})) \qquad (7)$$

Here, $p$ are the predictions, $t$ are the target labels, $i$ denote the data point, and $j$ denotes the class.

*D. System Structure*

The overall benchmarks and evaluation in section 4 illustrate the recent improvements in the DL architectures in computer vision. The result states that the recently investigated architectures mostly perform better in image recognition tasks. After completing the training procedure, the model is imported into the next module of the system. In this module, the end-user can dynamically detect the object. The end-user upholds the object to the web-camera, and the system automatically detects the object. The overall architecture of the proposed object classification system is demonstrated in Fig. 2.

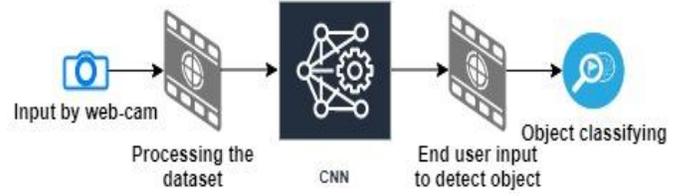

Fig. 2. The overall architecture of the proposed real-time object classification.

MobileNet [30] architecture gives the best accuracy for this study. Mobilenet architecture is built on depthwise separable convolutions, except for the first layer. The first layer is a full convolutional layer or dense layer. All layers are followed by batch normalization and ReLU non-linearity. However, the final layer is a fully connected layer without any non-linearity and feeds to the softmax for classification. For downsampling, stridden convolution is used for both depthwise convolutions as well as for the first fully convolutional layer. The total number of layers for MobileNet is 28 considering depthwise and pointwise convolution as separate layers.

IV. EVALUATION

In this section, firstly, the dataset condition and evaluation metrics are defined. Later, the empirical setup is explained. Finally, we present the evaluation with a detailed analysis.

*A. Dataset*

The proposed architecture is worked based on real-time user data. The user can fix the number of classes and the number of images for each category. To measure the efficiency of the model and measure the accuracy, we have used 500 images of 10 different classes such as mobile, pen, mouse, keyboard, speaker, paper, spray, book, pencil-box, and human. Web cameras have taken all the 5000 images before feeding into the architecture. Each image size is lies between 32 to 128 pixels. Some sample images from the dataset are shown in Fig. 3.

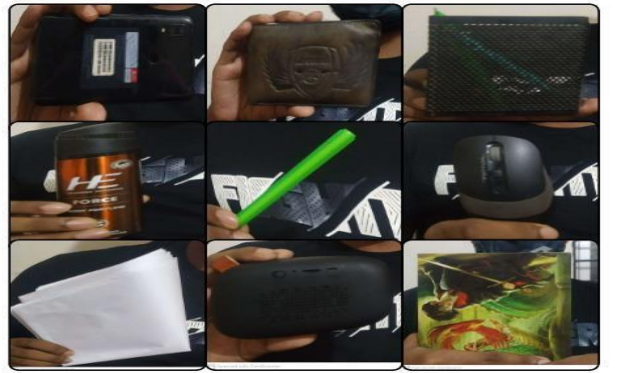

Fig. 3. The sample images of Smartphone, Money bag, Pencil-box, Spray, Pen, Mouse, Paper, Speaker, and Book from the dataset we used in this work of top left to bottom right.

## B. Evaluation Metric

We evaluated different baseline architecture using accuracy evaluation metrics which are based on the confusion matrix. The accuracy measurement shows how often the classifier gives the right prediction. The following equation calculates accuracy,

$$Accuracy = \frac{TP+TN}{TP+TN+FP+FN} \quad (8)$$

Where the four machine learning measures are true positive (T.P.), true negative (T.N.), false positive (F.P.), and false-negative (F.N.).

## C. Experimental Setup

JavaScript programming language is used for collecting data, pre-processing, experimenting, and evaluations of the model. The neural network architecture is implemented in TensorFlow.JS [34]. The dataset is created by webcam and processed using JavaScript.

## D. Evaluation Metric

For proper evaluation of every baseline architecture, each model was pre-trained on the ImageNet dataset. Each of the results showed in this paper is presented as a mean of 4 runs. Every model is trained with a limit of 50 epochs.

Table I illustrates the accuracy of the validation dataset of the different architectures. From this table, it can be said that nearly all of the models produce better accuracy. ResNet and VGG architecture hardly obtain the accuracy of 0.9, because of the overfitting in the training dataset. On the other hand, MobileNet achieves better compared to the trainable parameters available in the architecture. Among the various models, MobileNet reaches the highest accuracy of 0.97.

The results of every architecture are generated for the best weight found on the validation dataset. MobileNet performs better accuracy, among all the baseline architectures. On the other hand, Xception slightly falls off from the maximum result. From the overall observation, it can be indicated that MobileNet architecture performs better in object classification using low-quality image datasets.

TABLE I. THE TABLE SHOWS THE VALIDATION ACCURACY OF THE SEVEN BASELINE ARCHITECTURE OF CNNs. THE NUMBER OF PARAMETERS IN EACH BASELINE ARCHITECTURE IS ALSO SHOWN IN THE GIVEN TABLE.

| Model | Accuracy (imagenet-Dataset) | Accuracy (webcam-Dataset) |
|---|---|---|
| DenseNet169 [27] | 93.20 | 94.97 |
| InceptionV3 [28] | 93.70 | 94.95 |
| InceptionResNetV2 [29] | **95.30** | 94.95 |
| Xception [33] | 94.50 | 96.95 |
| ResNet50V2 [31] | 93.00 | 91.96 |
| VGG16 [32] | 90.10 | 90.53 |
| MobileNet [30] | 89.50 | **97.98** |

However, MobileNet is based on a streamlined structure that uses depthwise separable convolutions to build lightweight deep neural networks. Depthwise separable convolution is made up of two layers: depthwise convolutions and pointwise convolutions. Depthwise separable convolutions form factorized convolutions that factorize a standard convolution into a depthwise convolution, and a 1×1 convolution called a pointwise convolution. But the first layer is a full convolution. Besides, for both layers, MobileNet uses both Batchnorm and ReLU non-linearities.

Besides, MobileNet is specially designed for low-end devices like mobile and embedded vision applications. It is widely used on the minimum computational devices for its ideal size, speed, latency, and accuracy characteristics. In our work, the images we have taken by webcam is much lower resolution than normal image classification datasets. MobileNet performed significantly good results for this low-resolution image dataset.

Table I illustrates that MobileNet architecture does not perform well for the imagenet classification dataset but it acquires maximum accuracy in the low-resolution webcam image datasets. Hence, it is proved that MobileNet is superior for an embedded device and outperforms other popular CNN architecture at low-quality image classification.

## V. CONCLUSION & FUTURE SCOPE

This paper implements and tests a dynamic object classification method validated based on different image classifier baselines using low-quality webcam images. We practiced a transfer learning scheme to train and test different baseline architectures precisely. The paper focused on object classification based on CNN because of its strong ability of feature extraction and better advantage than traditional systems on real-time, accuracy, and adaptability. Further, we evaluate different architecture on a dataset that can be given by the end-user. We observe that famous baseline architectures like Xception, DenseNet, ResNet, etc. performs well for high-quality imagenet datasets but failed to perform expected result on low-quality image datasets. On the other hand, MobileNet performs best on low-quality images. However, it still has lots of chances for improvement. Advancing the architecture of CNN can minimize the loss of feature extraction. We strongly believe that this research work's contribution will pave the way for significant research on image processing approaches and will enhance the intelligence and the practicability of object classification based on CNN in future work.

## ACKNOWLEDGMENT

The authors would like to thank the Advanced Machine Learning (AML) lab for resource sharing and precious supports.